\documentstyle[colap]{article}

\newcommand{\fdrs}[2]
{~\fbox{\small $#1~\mid~#2$}~}

\newcommand{\lud}[3]
{ < 
      \left\{\begin{array}{l}
           #1
           \end{array}\right\} , 
      \left\{\begin{array}{l}
           #2
           \end{array}\right\} , 
      \left\{\begin{array}{l}
           #3
           \end{array}\right\}
  >       
}        

\newcounter{equationsave}          

{%
\begin{list}{(\theequation)}%
{%
\setcounter{equationsave}{\arabic{equation}}%
\usecounter{equation}
\setcounter{equation}{\arabic{equationsave}}
\setlength{\listparindent}{0pt}%
\def\makelabel##1{##1\hfil}
}%
\raggedright}
{\end{list}}

\title{\vspace{-0.5in}Compositional Semantics in Verbmobil}
\author{\begin{tabular}[c]{ccc}
	Johan Bos & Bj\"orn Gamb\"ack & Christian Lieske \\
        Yoshiki Mori & Manfred Pinkal & Karsten Worm
	\end{tabular}\\[2ex]
	Department of Computational Linguistics\\ 
	University of the Saarland \\
	Postfach 151150 \\
	D-66041 Saarbr\"ucken, Germany%
\thanks{\hspace*{4pt}This research was
funded by the German Federal Ministry of Education, Science,
Research, and Technology (BMBF) under grant number 01 IV 101 R.}
\\ 
	\verb!e-mail: vm@coli.uni-sb.de! \\
	~ \\
	{\it To appear in the Proceedings of COLING '96\/}}
\date{}

\begin{document}

\maketitle
\vspace{-0.5in}
\begin{abstract}

The paper discusses how compositional semantics is implemented in
the Verbmobil speech-to-speech translation system using
LUD, a description language for underspecified discourse
representation structures. The description language and its formal
interpretation in DRT are described as well as its implementation
together with the architecture of the system's entire syntactic-semantic
processing module. We show that a linguistically sound theory
and formalism can be properly implemented in a system with (near)
real-time requirements.

\end{abstract}

\section{Introduction}\label{Introduction}

Contemporary syntactic theories are normally unification-based and 
commonly aim at specifying as much as possible of the peculiarities of 
specific language constructions in the lexicon rather than in the ``traditional''
grammar rules. When doing semantic interpretation within such a 
framework, we want a formalism which allows for
\begin{itemize}
\item compositionality,
\item monotonicity, and
\item underspecification.
\end{itemize}

{\em Compositionality\/} may be defined rather strictly so that the 
interpretation of a phrase always should be the (logical) sum of the 
interpretations of its subphrases. A semantic formalism being compositional
in this strict sense would also trivially be {\em monotonic}, since no 
destructive changes would need to be undertaken while building the 
interpretation of a phrase from those of its subphrases.\footnote{More
formally, a semantic representation is monotonic iff the interpretation
of a category on the right side of a rule subsumes the interpretation of
the left side of the rule.}

However, compositionality is more commonly defined in a wider sense, 
allowing for other mappings from subphrase-to-phrase interpretation than 
the sum, as long as the mappings are such that the interpretation of the 
phrase still is a function of the interpretations of the subphrases. A 
common such mapping is to let the interpretation of the phrase be the 
interpretation of its (semantic) head modified by the interpretations of 
the adjuncts. If this modification is done by proper unification, the 
monotonicity of the formalism will still be guaranteed.

In many applications for Computational Linguistics, for example when 
doing semantically based translation --- as in Verbmobil, the German 
national spoken language translation project described in 
Section~\ref{Verbmobil} --- a complete interpretation of an utterance is not 
always needed or even desirable. Instead of trying to resolve ambiguities, 
for example the ones introduced by different possible scopings of 
quantifiers, the interpretation of the ambiguous part is left unresolved.
The semantic formalism of such a system should thus allow for the
{\em underspecification\/} of these unresolved ambiguities (but still
allow for them to be resolved in a monotonic way, of course). An
underspecified form representing an utterance is then the representation
of a {\em set\/} of meanings, all the possible interpretations of the
utterance.

The rest of the paper is structured as follows. Section~\ref{Verbmobil}
gives an overview of the Verbmobil Project. Section~\ref{LUD} introduces
LUD (description Language for Underspecified Discourse representations),
the semantic formalism we use. Section~\ref{Related} compares our approach
to that of others for similar tasks. The actual implementation is described
in Section~\ref{Implementation}, which also discusses coverage and points
to some areas of further research. Finally, Section~\ref{Conclusions} sums
up the previous discussion.

\section{The Verbmobil Project}\label{Verbmobil}

The project Verbmobil funded by the German Federal Ministry of Research and
Technology (BMBF) combines speech technology with machine translation
techniques in order to develop a system for translation in face-to-face
dialogues. The overall project is described in \cite{Wahlster:93};
in this section we will give a short overview of the key aspects.

The ambitious overall objective of the Verbmobil project is to produce a device
which will provide English translations of dialogues between German and
Japanese businessmen who only have a restricted active, but larger passive
knowledge of English. The domain is the scheduling of business appointments.
The major requirement is to provide translations as and when users need them,
and do so robustly and in (near) real-time.

In order to achieve this, the system is composed of time-limited processing 
components which on the source language (German or Japanese) side perform
speech recognition, syntactic, semantic and pragmatic analysis, as well as dialogue
management; transfer on a semantic level; and on the target language (English) side
generation and speech synthesis. When the users speak English, only keyword
spotting for the dialogue management is undertaken.

At any moment in the dialogue, a user may activate the Verbmobil device and start
speaking his/her native language. The speech recognition component then processes
the input and produces a word lattice representing the speech hypotheses and their
corresponding prosodic information. The parsing component processes the
lattice and assigns each well-formed path through it one or several syntactic
and (compositional) semantic representations. Ambiguities introduced by
these may be resolved by a resolution component.
The representations produced are then assigned dialogue acts and used to update
the model of the discourse, which in turn may be used by the speech recognizer to
choose the current language model. The transfer component takes the (possibly
resolved) semantic analysis of the input and builds a target language representation.
The generator then constructs the corresponding English expression. For
robustness, this deep-level processing strategy is complemented with a
shallow analysis-and-transfer component.

\section{Underspecified Representations}\label{LUD}

\subsection{Theoretical Background}\label{LUD-theory}

Since the Verbmobil domain is related to discourse rather than isolated
sentences, a variant of Kamp's Discourse Representation Theory, DRT
\cite{Kamp;Reyle:93} has been chosen as the model theoretic semantics.
However, to allow for underspecification of several linguistic phenomena,
we have chosen a formalism that is suited to represent underspecified
structures: LUD, a description language for underspecified discourse
representations \cite{Bos:95}. The basic idea is the one given in
Section~\ref{Introduction}, namely that natural language expressions
are not directly translated into Discourse Representation Structures
(DRSs), but into a representation that describes {\em several\/} DRSs. 

Representations in LUD have the following distinct features.
Firstly, all elementary semantic ``bits'' (conditions, entities, and events)
are uniquely labeled. This makes them easy to refer to and results in a
very powerful description language.
Secondly, meta variables over DRSs (which we call {\em holes\/}) allow
for the assignment of underspecified scope to a semantic operator.
Thirdly, a subordination relation on the set of holes and labels
constrains the number of interpretations of the LUD-representation
in the object language: DRSs.
 
\subsection{LUD-Representations}\label{LUD-Representations}

A LUD-representation $U$ is a triple 
\begin{eqnarray*}
<H_U,L_U,C_U>
\end{eqnarray*}
where $H_U$ is a set of holes (variables over labels), $L_U$ is a
set of labeled (LUD) conditions, and $C_U$ is a set of constraints.
A {\em plugging\/} is a bijective function from holes to labels. For
each plugging there is a corresponding DRS.
The syntax of LUD-conditions is formally defined as follows:

\begin{quote}
\begin{enumerate}
\item If $x$ is a discourse marker (i.e., entity or event),
	then $dm(x)$ is a LUD-condition;
\item If $R$ is a symbol for an $n$-place relation, $x_1,\dots,x_n$
	 are discourse markers,
	 then $pred(R,x_1,\dots,x_n)$ is a LUD-condition;
\item If $l$ is a label or hole for a LUD-condition,
	then $\lnot l$ is a LUD-condition;
\item If $l_1$ and $l_2$ are labels (or holes) for LUD-conditions, 
      then $l_1 \to l_2$, $l_1 \land l_2$ and $l_1 \lor l_2$ are LUD-conditions;
\item Nothing else is a LUD-condition.
\end{enumerate} 
\end{quote}

There are three types of constraints in LUD-representations. There is
{\em subordination\/} ($\leq$), {\em strict subordination\/} ($<$), 
and finally {\em presupposition\/} ($\alpha$). These constraints are
syntactically defined as:

\begin{quote}
If $l_1$, $l_2$ are labels, $h$ is a hole, then $l_1 \leq h$, $l_1 < l_2$ and
  $l_1~\alpha~l_2$ are LUD-constraints.  
\end{quote}

The interpretation of a LUD-representation is the interpretation
of {\bf top}, the label or hole of a LUD-representation for which there
exists no label that subordinates it.%
\footnote{The reader  interested in
a more detailed  discussion of the interpretation of underspecified
semantic representations is referred to \cite{Bos:95}.}

The interpretation function $I$ is a function from a labeled
condition to a DRS. This function is defined with respect to a
plugging $P$.
We represent a DRS as a box \fdrs{D}{C},
where $D$ is the set of discourse markers and $C$ is the set
of conditions. The mappings between LUD-conditions and DRSs are then
defined in (\ref{first-rule})-(\ref{last-rule}) where $l$ is a label or hole
and $\phi$ is a labeled condition.

\begin{eqnarray}
\lefteqn{I_P(l) =} \\\nonumber\label{first-rule}
	&&  I(\phi) \ {\rm iff} \ l:\phi \ \in L_U \\[6pt]
\lefteqn{I_P(l) =} \\\nonumber
	&&  I(P(l)) \  {\rm iff} \  l \in H_U\\[6pt]
\lefteqn{I(dm(x)) =} \\\nonumber
	&&  \left\{ \fdrs{x}{~} \right\} \\[6pt]
\lefteqn{I(pred(R,x_1,\dots,x_n)) =} \\\nonumber
	&&  \left\{ \fdrs{~}{R(x_1,\dots,x_n)} \right\} \\[6pt]
\lefteqn{I(l_1 \land l_2) =} \\\nonumber\label{merge-rule}
	&&  \left\{  K_1 \otimes K_2 \mid K_1 \in I(l_1)~\&~K_2 \in I(l_2) \right\} \\[6pt]
\lefteqn{I(l_1 \to l_2) =} \\\nonumber
	&&  \left\{ \fdrs{~}{K_1 \to K_2} \mid
                             K_1 \in I(l_1)~\&~K_2 \in I(l_2) \right\}\\[6pt]
\lefteqn{I(l_1 \lor l_2) =} \\\nonumber
	&&  \left\{ \fdrs{~}{K_1 \lor K_2} \mid
                             K_1 \in I(l_1)~\&~K_2 \in I(l_2) \right\} \\[6pt]
\lefteqn{I(\lnot l_1) =} \\\nonumber\label{last-rule}
	&&  \left\{ \fdrs{~}{\lnot K_1} \mid K_1 \in I(l_1) \right\}
\end{eqnarray}

In (\ref{merge-rule}) $\otimes$ is the merge operation, that takes two DRSs
$K_1$ and $K_2$ and returns a DRS which domain is the union of the set of
the domains of $K_1$ and $K_2$, and which conditions form the union of the
set of the conditions of $K_1$ and $K_2$. 

\subsection{Lexical Entries and Composition}\label{LUD-composition}

For building LUD-representations we use a lambda-operator
and functional application in order to compositionally 
combine simple LUD-representations to complex ones. In 
addition, we have two functions that help us to keep track of
the right labels. These are {\bf top}, as described above, and {\bf main}, 
the label of the semantic head of a LUD-representation. Further, we have
an operation that combines two LUD-representations into one:
$\oplus$ (merge for LUD-representations). Some sample lexical entries
for German

\begin{figure*}[htbp]
\begin{eqnarray*}
das\/:&&	\lambda P.
	\lud{~}{ l_i: dm(z)}{ l_i~\alpha~{\bf main}(P)}
	\oplus P(z)\\[6pt]
geht\/:&&	\lambda y.\lambda e.
	\lud{h_l}{l_i: pred(gehen,e),\\
                  l_j: pred(theme,e,y),\\
                  l_k: l_i \land l_j\\}
            {l_k \leq h_l}\\[6pt]
jeder\/:&&        \lambda P. \lambda Q.
        \lud{h_i}{ l_j: dm(x),\\
                   l_k: l_j \land {\bf main}(P),\\
                   l_l: l_k \to h_i}{ l_l \leq {\bf top}(Q),\\
                                      {\bf main}(Q) \leq h_i}
        \oplus P(z) \oplus Q(z)\\[6pt]
termin\/: &&   \lambda x.
       \lud{~}{l_i: termin(x)}{~}\\[6pt]
das~geht\/:&& \lambda e. \lud{h_0}
                                  {l_4: dm(z),\\
                                   l_5: pred(gehen,e),\\
                                   l_6: pred(theme,e,z),\\
                                   l_7: l_5 \land l_6\\} 
                                  {l_7 \leq h_0,\\ 
                                   l_4~\alpha_i~l_7\\}
\end{eqnarray*}
\caption{Lexical entries and a sample derivation in LUD}
\label{lexentries}
\end{figure*}

\section{Related Work}\label{Related}

The LUD representation is quite closely related to UDRSs,
underspecified DRSs \cite{Reyle:93}. The main difference is that the LUD
description language in principle is independent of the object language,
thus not only DRT, but also ordinary predicate logic, as well as a Dynamic
Predicate Logic \cite{Groenendijk;Stokhof:91} can be used as the object
language of LUD, as shown in \cite{Bos:95}. Compared to UDRS, LUD also
has a stronger descriptive power: Not DRSs, but the smallest possible
semantic components are uniquely labeled.

The Verbmobil system is a translation system built by some 30
different groups in three countries. The semantic formalism used on
the English generation side has been developed by CSLI, Stanford and
is called MRS, Minimal Recursion Semantics \cite{CopestakeEA:95}. The
deep-level syntactic and semantic German processing of Verbmobil is
also done along two parallel paths. The other path is developed by
IBM, Heidelberg and uses a variant of MRS, Underspecified Minimal
Recursion Semantics (UMRS) \cite{Egg;Lebeth:95}. All the three
formalisms LUD, MRS, and UMRS have in common
that they use a flat, neo-Davidsonian representation and allow for the
underspecification of functor-argument relations. In MRS,
this is done by unification of the relations with unresolved
dependencies.  This, however, results in structures which cannot be
further resolved.  In UMRS this is modified by
expressing the scoping possibilities directly as disjunctions. The
main difference between both types of MRSs and LUD is that the
interpretation of LUD in an object language other than ordinary
predicate logic is well defined, as described in
Section~\ref{LUD-Representations}.

The translation task of the SICS-SRI Bilingual Conversation Interpreter,
BCI \cite{AlshawiEA:91b} is quite similar to that of Verbmobil. The BCI
does translation at the level of Quasi-Logical Form, QLF which also is
a monotonic representation language for compositional semantics as discussed
in \cite{Alshawi;Crouch:92}. The QLF formalism incorporates a Davidsonian
approach to semantics, containing underspecified quantifiers and operators,
as well as `anaphoric terms' which stand for entities and relations to be
determined by reference resolution. In these respects, the basic ideas of
the QLF formalism are quite similar to LUD.

\section{Syntax--Semantics Interface and Implementation}\label{Implementation}

\subsection{Grammar}\label{Grammar}

The LUD semantic construction component has been implemented in the
grammar formalism TUG, Trace and Unification Grammar \cite{Block;Schachtl:92},
in a system called TrUG (in cooperation with Siemens AG, Munich, who provided
the German syntax and the TrUG system). TUG is a formalism that combines ideas
from Government and Binding theory, namely the use of traces, with
unification in order to account for, for example, the free
word order phenomena found in German.

\subsubsection{Syntax and Semantics}

A TUG grammar basically consists of PATR-II style context free
rules with feature annotations. Each syntactic rule gets annotated
with a semantic counterpart. In this way, syntactic derivation and
semantic construction are fully interleaved and semantics
can further constrain the possible readings of the input. 

In order to make our formalisation executable, we employ the
TrUG system, which compiles our rules into an efficient Tomita-style
parser. In addition TrUG incorporates sortal information, which is used
to rank parsing results.

Consider a simplified example of a syntactic rule annotated with a
semantic functor--argument application.

\begin{verbatim}
 s ---> np, vp |
    np:agr = vp:agr,
    lud_fun_arg(s,vp,np).
\end{verbatim}

In this example, a sentence {\tt s} consists of an {\tt np} and a {\tt
vp}. The first feature equation annotated to this rule says that the
value of the feature {\tt agr} (for agreement) of the {\tt np} equals
that of the respective feature value of the {\tt vp}.

\subsubsection{The Composition Process}

A category symbol like {\tt np} in the rule above also stands for the
entry node of its associated feature structure. This property is used
for the semantic counterpart of the rule: {\tt lud\_fun\_arg} is a call to
a semantic rule, a {\em macro\/} in the TUG notation,  which defines
functor--argument application. Since the macro gets the entry nodes of
the feature structures as arguments, all the information present in the
feature structures can be accessed within the macro which is defined as

\begin{verbatim}
 lud_fun_arg(Result,Fun,Arg) => 
    lud_context_equal(Fun,Result),
    context(Fun,FunContext),
    context(Arg,ArgContext),
    subcat(Result,ResultSc),
    subcat(Fun,[ArgContext|ResultSc]).
\end{verbatim}

The functor--argument application is based on the notion of the {\em
context\/} of a LUD-representation. The context of a
LUD-representation is a three-place structure consisting of the
LUD-representation's main label and top hole (as described in
Section~\ref{LUD-composition}) and its main instance, which is a discourse
marker or a lambda-bound variable. A LUD-representation also has a semantic
subcategorization list under the feature {\tt subcat} which performs the
same function as a $\lambda$--prefix. This list consists of the contexts
of the arguments a category is looking for.

The functor--argument application macro thus says the following. The
context of the result is the context of the functor. The functor is looking
for the argument as the first element on its {\tt subcat} list, while the
result's {\tt subcat} list is that of the functor minus the argument
(which has been bound in the rule). The binding of variables between
functor and argument takes place via the {\tt subcat} list, through
which a functor can access the main instance and the main label of its
arguments and state relations between them.

Note that the only relevant piece of information contained in a
LUD-representation for the purpose of composition is its
context. Its content in terms of semantic predicates is handled
differently. The predicates of a LUD-representation are stored in a
special slot provided for each category by the TrUG system. The
contents of this slot is handed up the tree from the daughters to
the mother completely monotonically. So the predicates introduced by
some lexical entry percolate up to the topmost node automatically.

These two restrictions, the use of only a LUD-representation's
context in composition and the monotonic percolation of semantic
predicates up the tree, make the system completely compositional in
the sense defined in Section~\ref{Introduction}.

\subsubsection{The lexicon}

To see how the composition interacts with the lexicon, consider the
following lexical macro defining the semantics of a transitive verb

\begin{verbatim}
 trans_verb_sem(Cat,Rel,[Role1,Role2]) => 
    basic_pred(Rel,Inst,L1),
    udef(Inst,L2),
    group([L1,L2,ArgL1,ArgL2],Main),
    leq(Main,Top),
    lud_context(Cat,Inst,Main,Top).
    role(Inst,Role1,Arg1,ArgL1),
    role(Inst,Role2,Arg2,ArgL2),
    subcat(Cat,[lud(Arg1,_,_),
                lud(Arg2,_,_)]).
\end{verbatim}

The macro states that a transitive verb introduces a basic
predicate of a certain relation with an instance and a label. The
instance is related to its two arguments by argument roles. The
arguments' instances are accessed via the verb's {\tt subcat} list
(and get bound during functor--argument application, cf. above). The
labels introduced are grouped together; the group label is the
main label of the LUD-representation, the instance its main
instance. Another property of the verb's semantics is that it
introduces the top hole of the sentence.

\subsection{Interfaces to Other Components}

As sketched in Section~\ref{Verbmobil}, our semantic construction
component delivers output to the components for semantic evaluation
and transfer. The paragraphs that follow describe the common interface
to these two components.

\subsubsection{Resolution of Underspecification}

Generating a scopally resolved LUD-re\-pre\-sen\-ta\-tion from an
underspecified one is the process which we referred to as plugging in
Section~\ref{LUD-Representations}.  It aims at making the possibly
ambiguous semantics captured by a LUD unique.  Obviously, purely
mathematical approaches for transforming the partial ordering encoded
in the \verb!leq! constraints into a total ordering may yield many results.

Fortunately, linguistic constraints allow us to reduce the effort that
has to be put into the computation of pluggings. An example is the
linguistic observation that a predicate that encodes sentence mood in
many cases modifies all of the remainder of the proposition for a
sentence. Thus, pluggings where the predicate for sentence mood is
subject to a \verb!leq! constraint should not be considered.  They would
result in a resolved structure expressing that the mood-predicate does
not have scope over the remaining proposition. This would be contrary
to the linguistic observation.

\subsubsection{Supplementary Information}

As a supplement to semantic predicates, our output contains various
kinds of additional information. This is caused by the overall
architecture of the Verbmobil system which does not provide for
fully-interconnected components. There is, e.g., no direct connection

between the speech recognizer and the component for semantic
evaluation. Thus, our component has to pipe certain kinds of information
(like prosodic values). Accordingly, our output consists of 
 ``Verbmobil Interface Terms'' (VITs), which differ slightly from the LUD-terms
described above mainly in that they include non-semantic information.

\subsection{Implementation Status}\label{Status}

Currently, the lexicon of the implemented system contains about 1400
entries (full forms) and the grammar consists of about 400 syntactic
rules, of which about 200 constitute a subgrammar for temporal
expressions. The system has been tested on three simplified dialogues
from a corpus of spoken language appointment scheduling dialogues
collected for the project and processes about $90$\% of the turns the
syntax can deal with.

The system is currently being extended to cover nine additional
dialogues from the corpus completely. The size of the lexicon will
then be about 2500 entries, which amounts to about 1700 lemmata.

\section{Conclusions}\label{Conclusions}

We have discussed the implementation of a compositional semantics
in the Verbmobil speech-to-speech translation system. The notions
of monotonicity and underspecification were discussed and LUD, a
description language for underspecified discourse representation
structures was introduced. As shown in Section~\ref{LUD},
the LUD description language has a well-defined interpretation in DRT.
Differently from Reyle's UDRSs, however, LUD assigns labels to the
minimal semantic element and may also be interpreted in other object
languages than DRT.

The key part of the paper, Section~\ref{Implementation}, showed how the
linguistically sound LUD formalism has been properly implemented in a
(near) real-time system. The implementation in Siemens' TUG grammar formalism
was described together with the architecture of the entire semantic processing
module of Verbmobil and its current coverage.

\section{Acknowledgements}\label{Acknowledgements}

We are gratefully emdebted to Scott McGlashan and CJ Rupp who both
worked on parts of the implementation. The results of the paper have
greatly benefitted from the cooperation with our other collegues in
Verbmobil, especially those at IBM and CSLI, as well as the ones working
on the modules closest to ours in the processing chain.

A number of people have contributed directly to parts of the work described
in the paper: Ronald Bieber, Hans-Ulrich Block, Michael Dorna, Manfred Gehrke,
Johannes Heinecke, Julia Heine, Daniela Kurz, Elsbeth Mastenbroek, Sebastian
Millies, Adam Przepiorkowski, Stefanie Schachtl, Michael Schiehlen, Feiyu Xu,
and several others.

\end{document}